\documentclass[runningheads]{llncs}
\usepackage{latexsym}
\usepackage{url}
\usepackage{multirow}
\usepackage{balance}
\usepackage{subfigure}
\usepackage{amsmath}
\usepackage{amssymb}
\usepackage{bm}
\usepackage{graphicx}
\usepackage{graphicx}
\usepackage{multicol}
\usepackage{multirow}
\usepackage{booktabs}
\usepackage{color}
\usepackage{transparent}
\usepackage{array}
\usepackage{balance}

\newcommand{\joint}{\text{JointKPE}}
\definecolor{darkpink}{rgb}{0.91, 0.33, 0.5}

\begin{document}
%
\title{Capturing Global Informativeness in Open Domain Keyphrase Extraction}
%

\author{Si Sun\inst{1}\thanks{Equal contributions.} \and
Zhenghao Liu\inst{2 \star} \and
Chenyan Xiong\inst{3} \and
Zhiyuan Liu\inst{4}\thanks{Corresponding authors.} \and
Jie Bao\inst{1 \star\star}}

\institute{Department of Electronic Engineering, Tsinghua University, China \and
Department of Computer Science and Technology, Northeastern University, China \and 
Microsoft Research, USA \and
Department of Computer Science and Technology, Tsinghua University, China \\
Institute for Artificial Intelligence, Tsinghua University, China \\
Beijing National Research Center for Information Science and Technology, China \\
\email{s-sun17@mails.tsinghua.edu.cn}; \email{liuzhenghao@cse.neu.edu.cn} \\
\email{chenyan.xiong@microsoft.com}; \email{\{liuzy, bao\}@tsinghua.edu.cn}
}


\maketitle

\begin{abstract}
Open-domain KeyPhrase Extraction (KPE) aims to extract keyphrases from documents without domain or quality restrictions, e.g., web pages with variant domains and qualities. Recently, neural methods have shown promising results in many KPE tasks due to their powerful capacity for modeling contextual semantics of the given documents. However, we empirically show that most neural KPE methods prefer to extract keyphrases with good phraseness, such as short and entity-style n-grams, instead of globally informative keyphrases from open-domain documents. This paper presents \joint{}, an open-domain KPE architecture built on pre-trained language models, which can capture both local phraseness and global informativeness when extracting keyphrases. \joint{} learns to rank keyphrases by estimating their informativeness in the entire document and is jointly trained on the keyphrase chunking task to guarantee the phraseness of keyphrase candidates. Experiments on two large KPE datasets with diverse domains, OpenKP and KP20k, demonstrate the effectiveness of \joint{} on different pre-trained variants in open-domain scenarios. Further analyses reveal the significant advantages of \joint{} in predicting long and non-entity keyphrases, which are challenging for previous neural KPE methods. Our code is publicly available at \url{https://github.com/thunlp/BERT-KPE}.

\end{abstract}

\section{Introduction}

Keyphrases that can provide a succinct summary of a document have exhibited their potential in improving many natural language processing (NLP) and information retrieval (IR) tasks~\cite{hasan2014automatic}, such as summarization~\cite{qazvinian2010citation}, recommendation~\cite{pudota2010automatic}, and document retrieval~\cite{jones1999phrasier}. High-quality keyphrases have shown two features, \textit{phraseness} and \textit{informativeness}~\cite{tomokiyo2003language}: \textit{phraseness} refers to the degree to which a sequence of words can serve as a complete semantic unit in the local contexts of documents; \textit{informativeness} indicates how well a text snippet captures the global topic or salient concept of the full document~\cite{liu2017phrase}. Many previous studies leverage the two features to improve the performance of KPE~\cite{Liu2010Automatic,El2014Scalable}.


With the development of neural networks, neural KPE methods have achieved convincing performance in extracting keyphrases for scientific publications~\cite{meng2017deep,sun2019divgraphpointer,mu2020keyphrase}. In recent years, many researchers have begun to adapt neural KPE to open-domain scenarios by considering diverse extraction domains with variant content qualities, such as the web KPE~\cite{xiong2019open}. Existing neural KPE methods often formulate keyphrase extraction as the tasks of word-level sequence tagging~\cite{zhang2018encoding}, n-gram level classification~\cite{xiong2019open}, and span-based extraction~\cite{mu2020keyphrase}. Despite their success, these neural methods seem to focus more on modeling the localized semantic features of keyphrases, which may cause them to prioritize local \textit{phraseness} rather than \textit{informativeness} of the global document when extracting keyphrases. As a consequence, they are inclined to extract keyphrases with semantic integrity from open-domain documents, such as short n-grams and head-ish entities, while long-tail phrases sometimes convey more pivotal information~\cite{dalvi2012analysis}.


In this paper, we propose \joint{} that servers open-domain keyphrase extraction scenarios. It can take both phraseness and informativeness into account when extracting keyphrases under a multi-task training architecture. \joint{} first resorts to powerful pre-trained language models to encode the documents and estimate the localized informativeness for all their n-grams. For the n-grams that have the same lexical strings but appear in various contexts, \joint{} further calculates their global informativeness scores in the full document. Lastly, \joint{} learns to rank these keyphrase candidates according to their global informative scores and jointly trains with the keyphrase chunking task~\cite{xiong2019open} to capture both local phraseness and global informativeness.



Experiments on two large-scale KPE datasets, OpenKP~\cite{xiong2019open} and KP20k~\cite{meng2017deep} with web pages and scientific papers, demonstrate \text{JointKPE}'s robust effectiveness with the widely used pre-trained model, BERT~\cite{devlin2019bert}, and its two pre-training variants, \text{SpanBERT}~\cite{joshi2020spanbert} and RoBERTa~\cite{liu2019roberta}. Our empirical analyses further indicate \text{JointKPE}'s advantages in predicting long keyphrases and non-entity keyphrases in open-domain scenarios.
\section{Related Work}

Automatic KeyPhrase Extraction (KPE) is concerned with automatically extracting a set of important and topical phrases from a document~\cite{turney2000learning,papagiannopoulou2020review}. Throughout the history of KPE, the earliest annotated keyphrase extraction corpora came from scientific domains, including technical reports and scientific literature~\cite{kim2013automatic,alzaidy2019bi,boudin2020keyphrase}. It is because scientific KPE corpora are easily curated whose keyphrases have already been provided by their authors. In addition to scientific corpora, some researchers have collected KPE corpora from the Internet and social media, e.g., news articles~\cite{wan2008collabrank} and live chats~\cite{kim2012extracting}, but these corpora are limited in size and lack diversity in terms of domains and topics. Recently, a large scale of open-domain KPE dataset named OpenKP~\cite{xiong2019open} has been released, which includes about one hundred thousand annotated web documents with a broad distribution of domains and topics. Based on the aforementioned corpora, extensive automatic KPE technologies have been proposed and examined. 

Existing KPE technologies can be categorized as unsupervised and supervised methods~\cite{papagiannopoulou2020review}. Unsupervised KPE methods are mainly based on statistical information~\cite{el2009kp,florescu2017new,campos2018text} and graph-based ranking algorithms~\cite{mihalcea2004textrank,liu2009clustering,rousseau2015main}. Supervised KPE methods usually formulate keyphrase extraction as phrase classification~\cite{witten2005kea,medelyan2009human,caragea2014citation} and learning-to-rank tasks~\cite{jiang2009ranking,Liu2010Automatic,zhang2017mike}. Thrived from sufficient well-annotated training data, supervised KPE methods, especially with neural networks, have significant performance advantages over state-of-the-art unsupervised methods in many KPE benchmarks~\cite{kim2013automatic,meng2017deep,xiong2019open}.

The earliest neural KPE methods often treat KPE as sequence-to-sequence learning~\cite{meng2017deep} and sequence labeling~\cite{zhang2018encoding} tasks, which adopts RNN-based encoder-decoder frameworks. The performance of these methods is limited by the shallow representations of textual data. Recently, Xiong et al.~\cite{xiong2019open} formulate KPE as an n-gram level keyphrase chunking task and propose BLING-KPE. It incorporates deep pre-trained representations into a convolutional transformer architecture to model n-gram representations. BLING-KPE achieves great improvement over previous methods. In addition, the recent work~\cite{ainslie2020etc} replaces the full self-attention of Transformers with local-global attention, which significantly boosts the KPE performance for long documents. More recently, Wang, et al.~\cite{wang2020incorporating} also show that incorporating multi-modal information in web pages, such as font, size, and DOM features, can bring further improvement for open-domain web KPE.

\section{Methodology}

Given a document $D$, \text{JointKPE} first extracts all keyphrase candidates $p$ from the document by enumerating its n-grams and utilizes a hierarchical architecture to model n-gram representations. Based on n-gram representations, \text{JointKPE} employs the informative ranking network to integrate the localized informativeness scores of multi-occurred phrases for estimating their global informativeness scores in the full document. During training, \text{JointKPE} is trained jointly with the keyphrase chunking task~\cite{xiong2019open} to better balance phraseness and informativeness.

\textbf{N-gram Representation.} \text{JointKPE} first leverages pre-trained language models, e.g., BERT~\cite{devlin2019bert}, to encode the document $D = \{w_1, \dots, w_{i},\dots,w_{n}\}$, and outputs a sequence of word embeddings $\bm{H}= \{\bm{h}_{1}, \dots, \bm{h}_{i}, \dots, \bm{h}_{n}\}$: 
\begin{equation}
\small
\bm{H} = \text{BERT}\{w_1, \dots, w_{i}, \dots, w_{n}\},
\end{equation}
where $\bm{h}_{i}$ is the word embedding of the $i$-th word $w_i$ in the document $D$.

To enumerate keyphrase candidates of the document $D$, the word embeddings are integrated into n-gram representations using a set of Convolutional Neural Networks (CNNs) since keyphrases usually appear in the form of n-grams. The representation of the $i$-th $k$-gram $c_i^k=w_{i:i+k-1}$ is calculated as:
\begin{equation}
\small
\bm{g}^{k}_{i} = \text{CNN}^{k}\{\bm{h}_{i}, \dots, \bm{h}_{i+k-1}\},
\end{equation}
where each $k$-gram is composed by its corresponding $\text{CNN}^{k}$ with the window size of $k$ ($1 \leq k \leq K$). $K$ is the maximum length of the extracted n-grams.

\textbf{Informative Ranking.} To estimate the informativeness for n-gram $c_i^k$ in local contexts, \text{JointKPE} takes a feedforward layer to project its context-specific representation $\bm{g}^{k}_{i}$ to a quantifiable score:
\begin{equation}
\small
f(c_i^k, D) = \text{Feedforward}(\bm{g}^{k}_{i}).
\end{equation}

We further compute global informativeness scores for those phrases that appear multiple time in different contexts of the document. Specifically, let phrase $p^k$ be a multi-occurred phrase of length $k$ in the document $D$. This phrase occurs in different contexts $\{c_j^k, \dots, c_l^k, \dots, c_m^k\}$ of the document, which thus leads to diverse localized informativeness scores $\{f(c_j^k, D), \dots, f(c_l^k, D), \dots, f(c_m^k, D)\}$. For this multi-occurred phrase, \text{JointKPE} applies max-pooling upon its localized informativeness scores to determine the global informativeness score $f^*(p^k, D)$:
\begin{equation}
\small
f^*(p^k, D) = \max \{f(c_j^k, D), \dots, f(c_l^k, D), \dots, f(c_m^k, D)\}.
\end{equation}

After estimating the global informativeness scores for all phrases in the document $D$, \text{JointKPE} can learn to rank these phrases in the document level based on their global informativeness scores using the pairwise ranking loss:
\begin{equation}
\small
L_\text{Rank} = \sum\limits_{p_+, p_- \in D}\text{max}(0, 1 - f^*(p_+, D) + f^*(p_-, D)),
\end{equation}
where the ranking loss $L_\text{Rank}$ enforces \text{JointKPE} to rank keyphrases $p_+$ ahead of non-keyphrases $p_-$ within the same document $D$.

\textbf{Keyphrase Chunking.} To enhance the phraseness measurement at the n-gram level, \text{JointKPE} combines the keyphrase chunking task~\cite{xiong2019open} to directly learn to predict the keyphrase probability of n-grams by optimizing the binary classification loss $L_\text{Chunk}$:
\begin{equation}
\small
L_\text{Chunk} = \text{CrossEntropy}(P(c_i^k=y_i^{k})),
\end{equation}
where $y_i^{k}$ is the binary label, which denotes whether the n-gram $c_i^k$ exactly matches the string of a keyphrase annotated in the document.

\textbf{Multi-Task Training.} The ultimate training objective of \text{JointKPE} is to minimize the linear combination of the informative ranking loss $L_\text{Rank}$ and the keyphrase chunking loss $L_\text{Chunk}$:
\begin{equation}
\small
L = L_\text{Rank} + L_\text{Chunk}. \label{eq.multi-loss}
\end{equation}

At the inference stage, the top-ranking phrases with the highest global informativeness scores are predicted as keyphrases for the given document.

\begin{table}
\caption{\label{tab:overall}Overall accuracy of keyphrase extraction. All scores of JointKPE on OpenKP are from the official leaderboard with a blind test. The baseline evaluation results on KP20k are obtained from corresponding papers, and the asterisked baselines* are our implementations. Bold \textbf{F1@3} and \textbf{F1@5} are main evaluation metrics of OpenKP and KP20k, respectively.}
\resizebox{\textwidth}{!}{
\begin{tabular}{l||c c c|c c c|c c c||c c} \hline \multirow{2}{2.25cm}{\textbf{Methods}} & \multicolumn{9}{c||}{\textbf{OpenKP}} & \multicolumn{2}{c}{\textbf{KP20k}} \\ 
\cline{2-12} & F1@1 & \textbf{F1@3} & F1@5 & P@1 & P@3 & P@5 & R@1 & R@3 & R@5 & \textbf{F1@5} & F1@10 \\ \hline 
\multicolumn{12}{l}{\textbf{Baselines}} \\ \hline
TFIDF~\cite{xiong2019open,mu2020keyphrase} & 0.196 & 0.223 & 0.196 & 0.283 & 0.184 & 0.137 & 0.150 & 0.284 & 0.347 & 0.108 & 0.134 \\
TextRank~\cite{xiong2019open,mu2020keyphrase} & 0.054 & 0.076 & 0.079 & 0.077 & 0.062 & 0.055 & 0.041 & 0.098 & 0.142 & 0.180 & 0.150\\
Maui~\cite{mu2020keyphrase} & n.a. & n.a. & n.a. & n.a. & n.a. & n.a. & n.a. & n.a. & n.a. & 0.273 & 0.240 \\
PROD~\cite{xiong2019open} & 0.245 & 0.236 & 0.188 & 0.353 & 0.195 & 0.131 & 0.188 & 0.299 & 0.331 & n.a. & n.a. \\
\hline
CopyRNN~\cite{meng2017deep} & 0.217 & 0.237 & 0.210 & 0.288 & 0.185 & 0.141 & 0.174 & 0.331 & 0.413 & 0.327 & 0.278 \\
CDKGEN~\cite{diao2020keyphrase} & n.a. & n.a. & n.a. & n.a. & n.a. & n.a. & n.a. & n.a. & n.a. & 0.381 & 0.324 \\
BLING-KPE~\cite{xiong2019open} & 0.267 & 0.292 & 0.209 & 0.397 & 0.249 & 0.149 & 0.215 & 0.391 & 0.391 & n.a. & n.a. \\ \hline
SKE-Base-Cls (BERT-Base)~\cite{mu2020keyphrase} & n.a. & n.a. & n.a. & n.a. & n.a. & n.a. & n.a. & n.a. & n.a. & 0.386 & 0.326 \\
Span Extraction (BERT-Base)* & 0.318 & 0.332 & 0.289 & 0.476 & 0.285 & 0.209 & 0.253 & 0.436 & 0.521 & 0.393 & 0.325  \\
Sequence Tagging (BERT-Base)* & \textbf{0.321} & \textbf{0.361} & \textbf{0.314} & \textbf{0.484} & \textbf{0.312} & \textbf{0.227} & \textbf{0.255} & \textbf{0.469} & \textbf{0.563} & \textbf{0.407} & \textbf{0.335}  \\ \hline
\multicolumn{12}{l}{\textbf{Our Methods}} \\ \hline
\text{JointKPE (BERT-Base)} & \textbf{0.349} & \textbf{0.376} & \textbf{0.325 }& \textbf{0.521} & \textbf{0.324} & \textbf{0.235} & \textbf{0.280} & \textbf{0.491} & \textbf{0.583} & 0.411 & 0.338 \\
Only Informative Ranking  & 0.342 & 0.374 & \textbf{0.325} & 0.513 & 0.323 & \textbf{0.235} & 0.273 & 0.489 & 0.582 & \textbf{0.413} & \textbf{0.340} \\ 
Only Keyphrase Chunking & 0.340 & 0.356 & 0.311 & 0.511 & 0.306 & 0.225 & 0.271 & 0.464 & 0.558 &  0.412 & 0.337 \\ \hline
\text{JointKPE (SpanBERT-Base)} & \textbf{0.359} & \textbf{0.385} & \textbf{0.336} & \textbf{0.535} & \textbf{0.331} & \textbf{0.243} & \textbf{0.288} & \textbf{0.504} & \textbf{0.603} & \textbf{0.416} & \textbf{0.340} \\
Only Informative Ranking  & 0.355 & 0.380 & 0.331 & 0.530 & 0.327 & 0.240 & 0.284 & 0.497 & 0.593 & 0.412 & 0.338 \\
Only Keyphrase Chunking & 0.348 & 0.372 & 0.324 & 0.523 & 0.321 & 0.235 & 0.278 & 0.486 & 0.581 & 0.411 & 0.338 \\ \hline
\text{JointKPE (RoBERTa-Base)} & \textbf{0.364} & \textbf{0.391} & \textbf{0.338} & \textbf{0.543} & \textbf{0.337} & \textbf{0.245} & \textbf{0.291} & \textbf{0.511} & \textbf{0.605} & \textbf{0.419} & \textbf{0.344} \\
Only Informative Ranking  & 0.361 & 0.390 & 0.337 & 0.538 & \textbf{0.337} & 0.244 & 0.290 & 0.509 & 0.604 & 0.417 & 0.343 \\
Only Keyphrase Chunking & 0.355 & 0.373 & 0.324 & 0.533 & 0.322 & 0.235 & 0.283 & 0.486 & 0.581 & 0.408 & 0.337 \\ \hline
\end{tabular}}
\end{table}


\section{Experimental Methodology}

This section introduces our experimental settings, including datasets, evaluation metrics, baselines, and implementation details.

\textbf{Dataset.} Two large-scale KPE datasets are used in our experiments. They are OpenKP~\cite{xiong2019open} and KP20k~\cite{meng2017deep}. OpenKP is an open-domain keyphrase extraction dataset with various domains and topics, which contains over 150,000 real-world web documents along with the most relevant keyphrases generated by expert annotation. We consider OpenKP as our main benchmark and follow its official split of training (134k documents), development (6.6k), and testing (6.6k) sets. KP20k is a scientific KPE dataset with the computer science domain, which consists of over 550,000 articles with keyphrases assigned by their authors. We follow the original work's partition of training (528K documents), development (20K), and testing (20K) sets~\cite{meng2017deep}.

\textbf{Evaluation Metrics.} Precision (P), Recall (R), and F-measure (F1) of the top $N$ keyphrase predictions are used for evaluating the performance of KPE methods. Following prior research~\cite{xiong2019open,meng2017deep}, we utilize $N = \{1, 3, 5\}$ on OpenKP and $N = \{5, 10\}$ on KP20k, and consider F1@3 and F1@5 as the main metrics for OpenKP and KP20k, respectively. For KP20k, we use Porter Stemmer~\cite{porter1980algorithm} to determine the match of two phrases, which is consistent with prior work~\cite{meng2017deep}.


\textbf{Baselines.} Two groups of baselines are compared in our experiments, including \textit{Traditional KPE baselines} and \textit{Neural KPE baselines}.

\textit{Traditional KPE baselines.} Keeping consistent with previous work~\cite{meng2017deep,xiong2019open,mu2020keyphrase}, we compare \joint{} with four traditional KPE methods. They are two popular unsupervised KPE methods, TFIDF and TextRank~\cite{mihalcea2004textrank}, and two feature-based KPE systems that perform well on OpenKP and KP20k, named PROD~\cite{xiong2019open} and Maui~\cite{medelyan2009human}. For these baselines, we take their reported performance.


\textit{Neural KPE baselines.} We also compare our method with six neural KPE baselines: the previous state-of-the-art method on OpenKP, \text{BLING-KPE}~\cite{xiong2019open} and three advanced neural KPE methods on KP20k, including \text{CopyRNN}~\cite{meng2017deep}, \text{CDKGEN}~\cite{diao2020keyphrase} and \text{SKE-Base-Cls}~\cite{mu2020keyphrase}, and two BERT-based KPE methods we have reconstructed, including Span Extraction and Sequence Tagging. They formulate KPE as the tasks of span extraction and sequence tagging, respectively. 

\textbf{Implementation Details.} The base versions of BERT~\cite{devlin2019bert}, \text{SpanBERT}~\cite{joshi2020spanbert} and RoBERTa~\cite{liu2019roberta}, initialized from their pre-trained weights, are used to implement \joint{}. All our methods are optimized using Adam with a 5e-5 learning rate, 10\% warm-up proportion, and 64 batch size. We set the maximum sequence length to 512 and simply keep the weights of the two training losses (Eq.~\ref{eq.multi-loss}) to be the same. Following previous work~\cite{meng2017deep,xiong2019open}, the maximum phrase length is set to 5 ($K=5$). The training used two Tesla T4 GPUs and took about 25 hours on three epochs. Our implementations are based on PyTorch-Transformers~\cite{wolf2020transformers}. 

\section{Results and Analyses}

In this section, we present the evaluation results of \joint{} and conduct a series of analyses and case studies to study its effectiveness.

\begin{figure}[t]
\centering 
\subfigure[Length Performance.]{
\includegraphics[height=4.4cm]{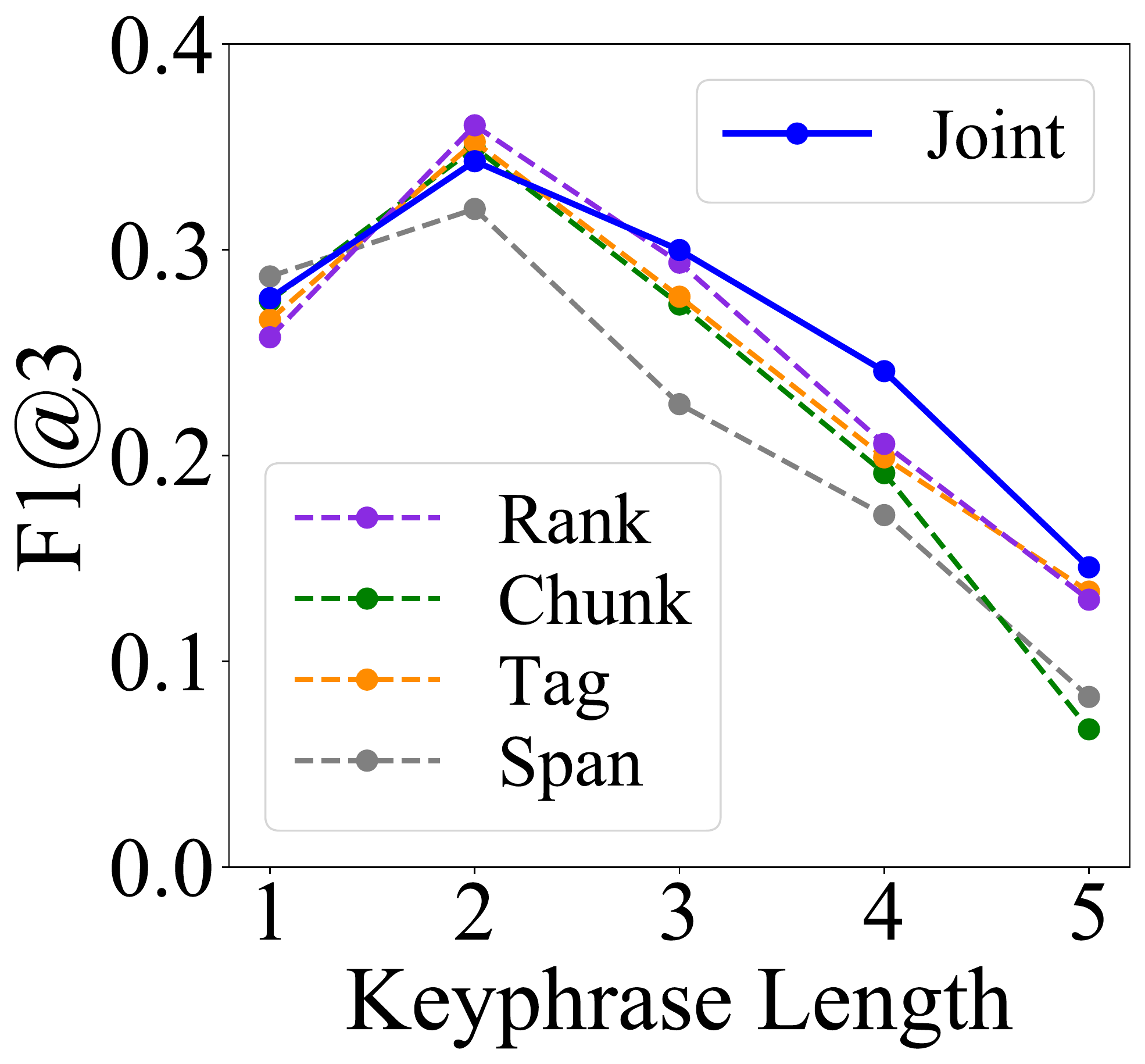}
\label{tab:length_performance}
}
\subfigure[Length Distribution.]{
\includegraphics[height=4.4cm]{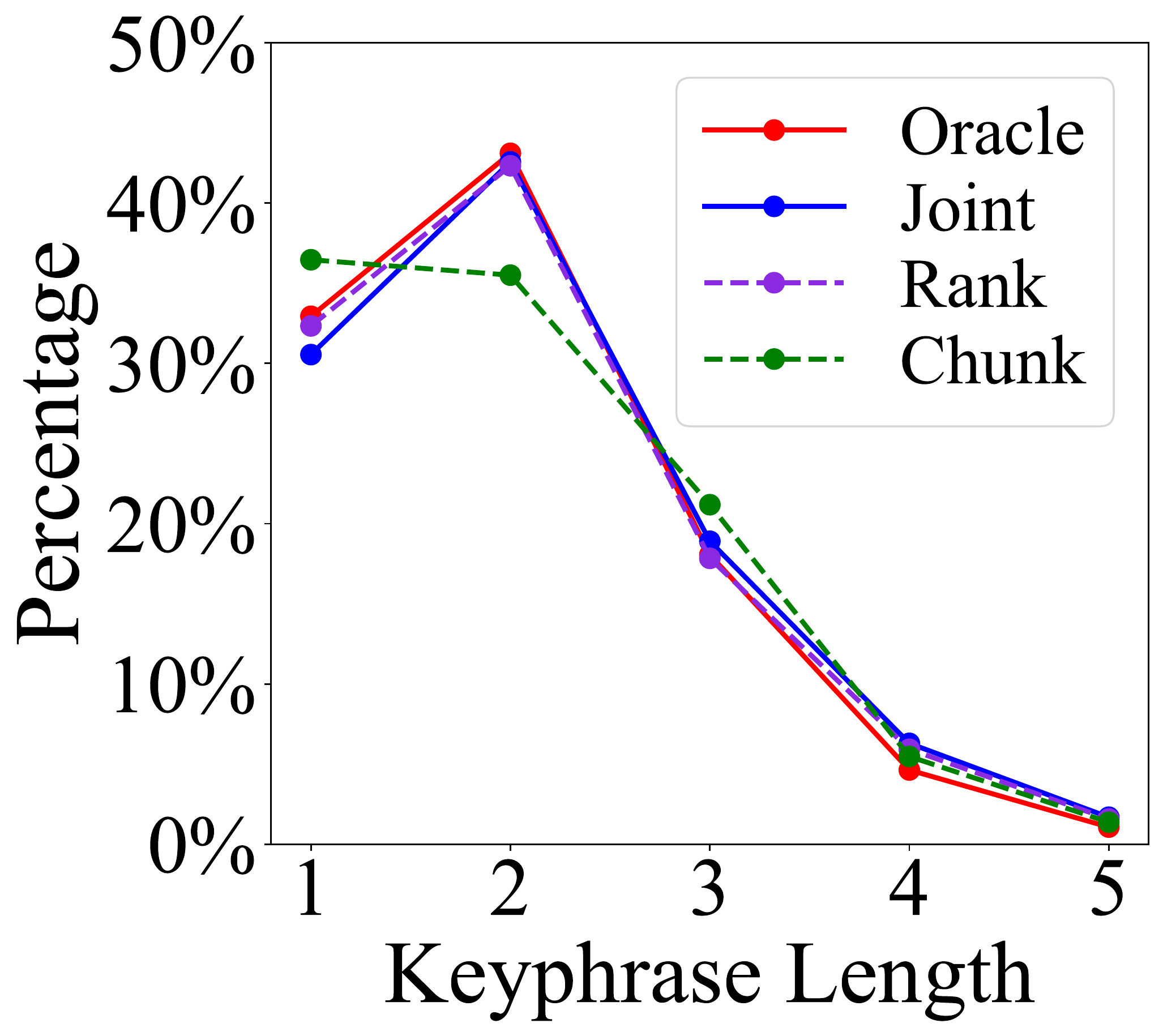}
\label{tab:length_distribution}
} 
\caption{\label{tab:length}Keyphrase length analyses. Figure~\ref{tab:length_performance} exhibits the F1@3 score of different neural KPE methods on OpenKP at different keyphrase lengths. Figure~\ref{tab:length_distribution} illustrates the length distribution of keyphrases predicted by \text{JointKPE} (Joint) and its two ablated versions: only informative ranking (Rank) and only keyphrase chunking (Chunk). Methods of \text{Sequence Tagging} and \text{Span Extraction} are abbreviated as Tag and Span, respectively.}
\end{figure}


\subsection{Overall Accuracy}
The evaluation results of \joint{} and baselines are shown in Table~\ref{tab:overall}.

Overall, \text{\joint{}} outperforms all baselines on all evaluation metrics stably on both open-domain and scientific KPE benchmarks. Compared to the best feature-based KPE systems, PROD~\cite{xiong2019open} and Maui~\cite{medelyan2009human}, \joint{} outperforms them by large margins. \joint{} also outperforms the strong neural baselines, \text{BLING-KPE}~\cite{xiong2019open} and \text{CDKGEN}~\cite{diao2020keyphrase}, the previous state-of-the-art on OpenKP and KP20k; their F1@3 and F1@5 are improved by 22.8\% and 7.9\% respectively. These results demonstrate \joint{}'s effectiveness.

Moreover, even with the same pre-trained model, \joint{} still achieves better performance than those neural KPE methods mainly based on localized features, e.g., \text{SKE-Base-Cls}~\cite{mu2020keyphrase}, Span Extraction, and Sequence Tagging. These results reveal that the effectiveness of \joint{}, in addition to the benefits gained from the pre-trained model, also stems from its ability to combine local and global features when extracting keyphrases.

As shown in the ablation results of \text{\joint{}}, without either informative ranking or keyphrase chunking, its accuracy declines, especially in the absence of the informative ranking on the open-domain OpenKP dataset. The result indicates the benefit of multi-task learning and the key role of capturing global informativeness in open-domain keyphrase extraction scenarios. 

Besides, \text{\joint{}}'s effectiveness can be further improved with initialized from \text{SpanBERT}~\cite{joshi2020spanbert} and \text{RoBERTa}~\cite{liu2019roberta}, the two BERT variants with updated pre-training strategies, which shows \joint{}'s ability to utilize the advantages of better pre-trained models. Furthermore, we observe that our method obtains more significant improvements on RoBERTa due to the higher enhancement of informative ranking performance, and the improvements brought by SpanBERT and RoBERTa to the keyphrase chunking task are relatively close\footnote{Due to space limitations, more baseline results based on \text{SpanBERT} and \text{RoBERTa} can be found in~\url{https://github.com/thunlp/BERT-KPE}.}.

\setcounter{table}{1}
\begin{table}[t]
\small
\centering
\caption{\label{tab:type_examples} Examples of entity and non-entity keyphrases in the OpenKP dataset.}
\begin{tabular}{l|l}  \hline
\textbf{Entity Keyphrase} & \textbf{Entity Type} \\ \hline
facebook & business.brand \\
the happy wanderer & music.composition \\
health and human services & government.government\_agency \\
\hline \hline
\textbf{Non-Entity Keyphrase} & \textbf{Type} \\ \hline 
join a world & verb phrase \\
care tenderly & adverbial phrase \\
beautiful clothing & adjective phrase \\
opposite a monkey & prepositional phrase \\
i have seen the light & sentence snippet \\
firing wice with double click & complex phrase \\
landmarks and historic buildings & composition phrase \\
\hline
\end{tabular}
\end{table}


\subsection{Performance w.r.t. Keyphrase Lengths}

One challenge in open-domain KPE is the long-keyphrase extraction scenario~\cite{xiong2019open}. In this experiment, we compare the behavior and effectiveness of \joint{} and other BERT-based KPE methods in extracting keyphrases of different lengths, using the open-domain OpenKP dataset.

As shown in Figure~\ref{tab:length_performance}, all neural KPE methods have better extraction performance for short keyphrases than longer ones. Nevertheless, compared with other methods, \text{\joint{}} has a more stable extraction ability for keyphrases of different lengths and performs best in predicting longer keyphrases (length $\geq$ 4). Notably, the F1@3 score of \text{\joint{}} is 17\% higher than its two ablated versions, Rank (only informative ranking) and Chunk (only keyphrase chunking), at the keyphrase length of 4. These results reveal that \joint{}'s joint modeling of phraseness and informativeness helps alleviate the long-keyphrase challenge in open-domain extraction scenarios.

Figure~\ref{tab:length_distribution} further illustrates the length distribution of keyphrases predicted by \text{\joint{}} and its two ablated versions, Chunk and Rank. The ablated model trained only with the keyphrase chunking task (Chunk)~\cite{xiong2019open} is inclined to predict more single-word phrases. By contrast, the keyphrase length distributions predicted by our full model (Joint) and its ranking-only version (Rank) are more consistent with the ground-truth length distribution. The result further shows that globally capturing informativeness can guide the neural extractor to better determine the boundaries of truth keyphrases.

\begin{figure}[t]
\centering 
\subfigure[Entity Performance.]{
\includegraphics[height=3.95cm]{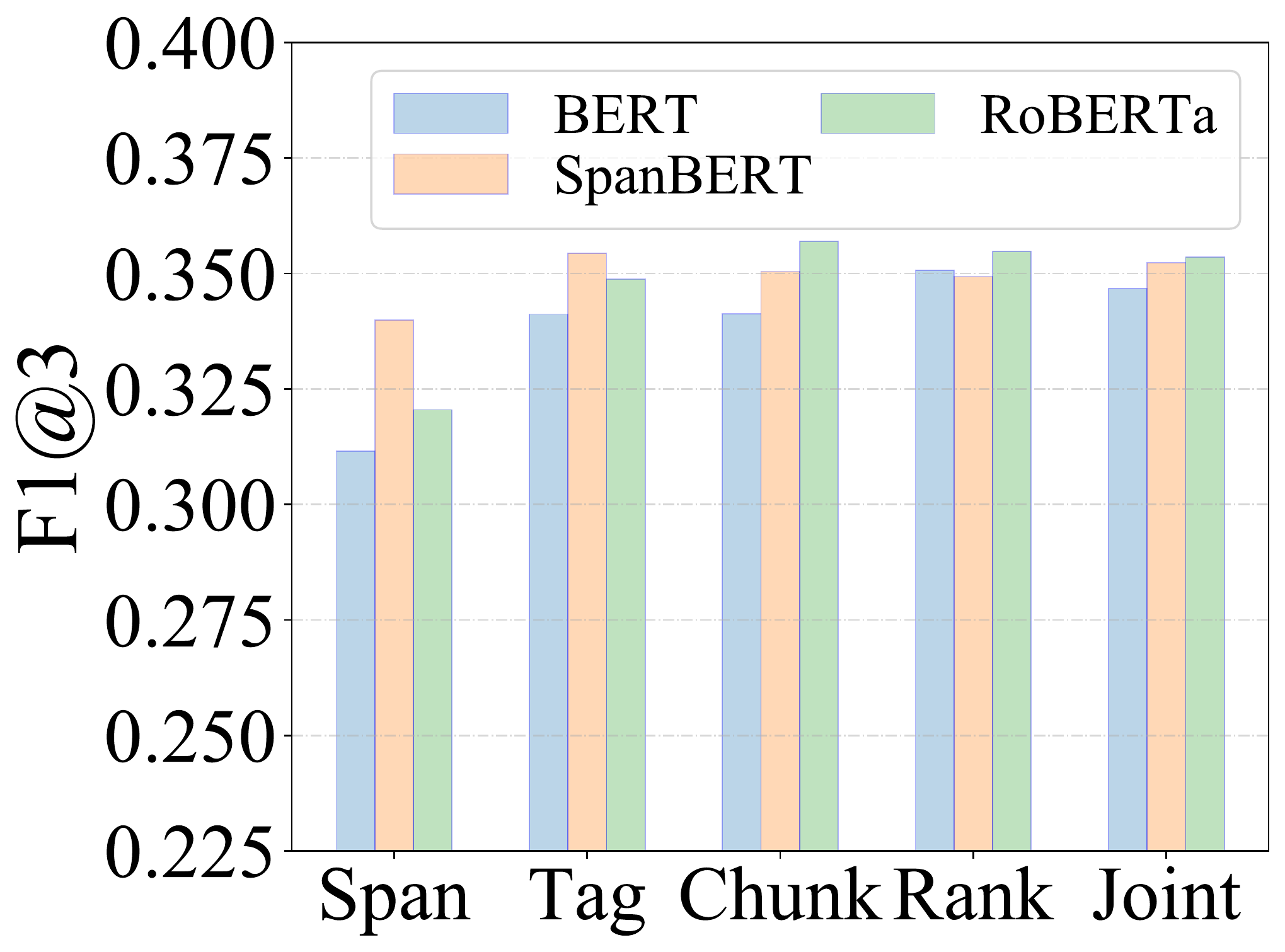}
\label{tab:type_b}
}
\subfigure[Non-Entity Performance.]{
\includegraphics[height=3.95cm]{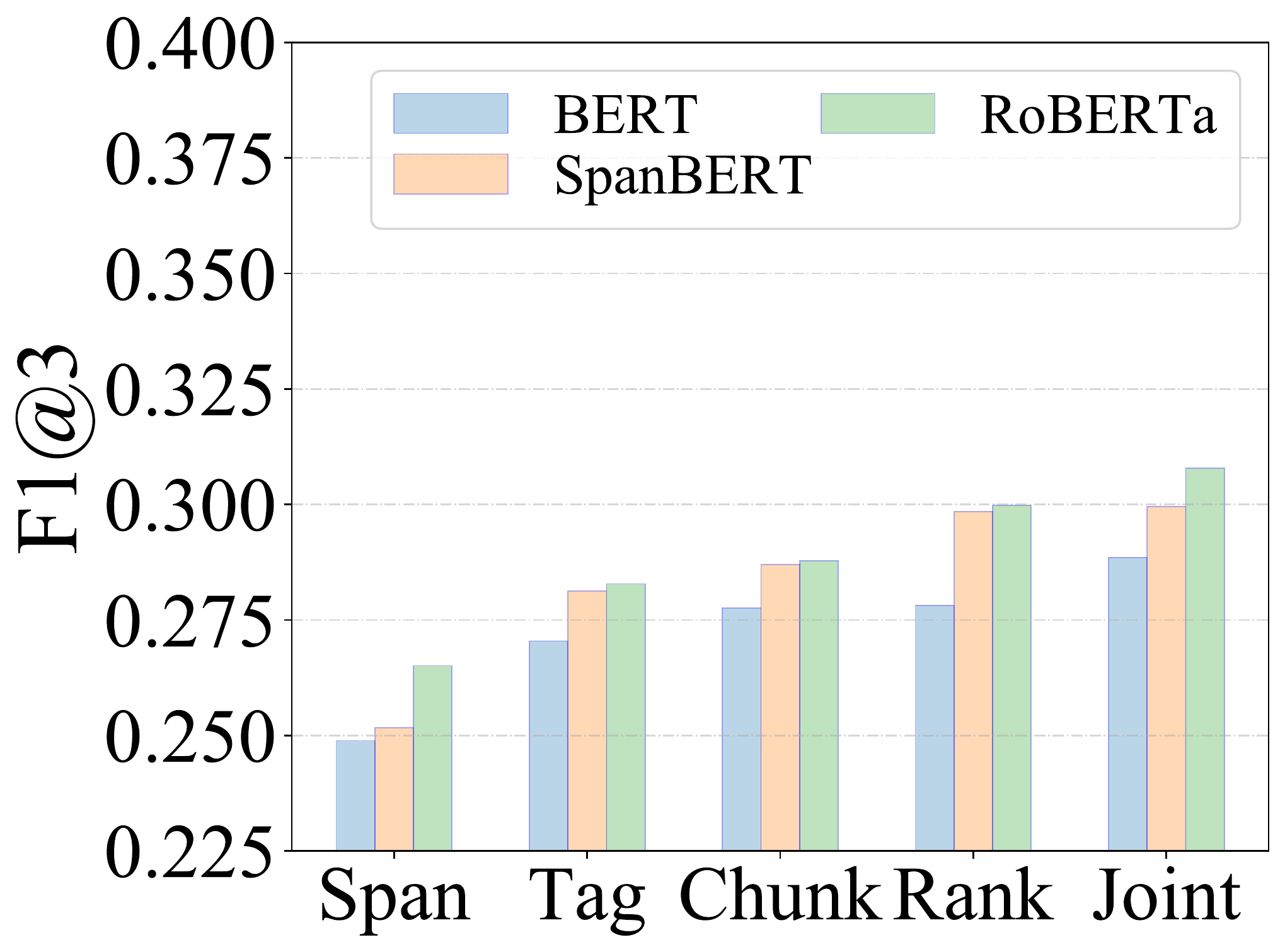}
\label{tab:type_c}
}
\caption{\label{tab:type}Keyphrase type analyses. Figure~\ref{tab:type_b} and Figure~\ref{tab:type_c} show the extraction performance for entity keyphrases and non-entity keyphrases on the OpenKP dataset, respectively. Five BERT-based KPE models, Span Extraction (Span), Sequence Tagging (Tag), the two ablated versions (Chunk and Rank), and the full model of JointKPE (Joint) are compared.}
\end{figure}


\subsection{Performance w.r.t. Keyphrase Types}

One of the main differences between keyphrases and entities is that not all keyphrases are knowledge graph entities, especially in open-domain KPE scenarios. From our observation of OpenKP, about 40\% of ground-truth keyphrases are not entities, which accounts for a significant portion. As examples exhibited in Table~\ref{tab:type_examples}, non-entity keyphrases often have higher-variant forms and may be more complicated to identify from the contexts. This experiment studies \text{\joint{}}'s effectiveness for non-entity keyphrases, where its two ablated versions and another two BERT-based KPE baselines are used as the comparison.

Figure~\ref{tab:type_b} and Figure~\ref{tab:type_c} illustrate the performance of different BERT-based KPE methods on entity keyphrases and non-entity keyphrases, where entities are identified by the CMNS linker~\cite{Xiong2016BOE}. Not surprisingly, all KPE methods with pre-trained models perform well in extracting entity keyphrases. Entity-style keyphrases with distinct and independent existence are easily recognized from the contexts. Also, pre-trained models might already capture the entity information and simple layers upon them can effectively identify entity keyphrases. 

\begin{table*}[t]
\small
\centering
\caption{\label{tab:cases}Prediction cases of long and non-entity keyphrases targeting OpenKP. Ground-truth keyphrases are \underline{\textbf{underlined}} in the document. The top three predictions of \text{JointKPE} (Joint) and its two ablated versions (\text{Rank} and \text{Chunk}) are exhibited. All three methods extracted \textbf{\textcolor{blue}{blue keyphrases}}, while \textbf{\textcolor{red}{red keyphrases}} were only predicted by \text{JointKPE}'s full version.}
\begin{tabular}{p{1.0\columnwidth}} 
\hline
\textbf{Long Keyphrase Cases} \\ \hline
Document: ... \underline{\textbf{walter lagraves}} on \underline{\textbf{mosquito control board race}} i will not vote for by \underline{\textbf{walter lagraves}} dear editor ralph depalma is a candidate for the board of the florida keys mosquito control authority mc lets think about ralph depalmas record ... \\

\textbf{@Joint:} \textbf{\textcolor{blue}{walter lagraves}}; \textbf{\textcolor{red}{mosquito control board race}}; mosquito control \\
\textbf{@Rank:} mosquito control; \textbf{\textcolor{blue}{walter lagraves}}; ralph depalma \\
\textbf{@Chunk:} \textbf{\textcolor{blue}{walter lagraves}}; ralph depalma; mosquito control board \\

\hline
Document: ... you can use in the know education central \underline{\textbf{compliance corner}} case notes enforcement actions ... new laws may have been enacted subsequently compliance topic of the month \underline{\textbf{surplus lines export eligibility}} the department has received numerous questions from agents seeking guidance on how to comply with ... \\

\textbf{@Joint:} \textbf{\textcolor{blue}{compliance corner}}; surplus lines; \textbf{\textcolor{red}{surplus lines export eligibility}} \\
\textbf{@Rank:} \textbf{\textcolor{blue}{compliance corner}}; surplus lines; insurance business \\
\textbf{@Chunk:} \textbf{\textcolor{blue}{compliance corner}}; export eligibility; surplus lines \\

\hline
\hline

\textbf{Non-Entity Keyphrase Cases} \\ \hline
Document: how to transfer \underline{\textbf{dv to laptop}} without having a \underline{\textbf{firewire port}} we will be using pinnacle moviebox update it does work ... here i compare the quality of a video transferred to the laptop via \underline{\textbf{firewire port}} vs pinnacle moviebox ...\\

\textbf{@Joint:} \textbf{\textcolor{blue}{firewire port}}, \textbf{\textcolor{red}{dv to laptop}}, dv \\
\textbf{@Rank:} \textbf{\textcolor{blue}{firewire port}}, laptop, dv \\
\textbf{@Chunk:} \textbf{\textcolor{blue}{firewire port}}, dv, laptop \\
\hline

Document: ... \underline{\textbf{north country realty}} welcome we deal in \underline{\textbf{the beauty of the north}} we look forward to working with you soon whether you are selling or buying north country property \underline{\textbf{north country realty}} is here to be of assistance to you we will gladly offer references from past clients and customers we have served ... \\

\textbf{@Joint:} \textbf{\textcolor{blue}{north country realty}}; oscoda; \textbf{\textcolor{red}{beauty of the north}} \\
\textbf{@Rank:} \textbf{\textcolor{blue}{north country realty}}; oscoda; north country property \\
\textbf{@Chunk:} \textbf{\textcolor{blue}{north country realty}}; north country property; real estate \\

\hline
\end{tabular}
\end{table*}


Nevertheless, the accuracy of all methods drops dramatically when extracting non-entity keyphrases, which evidently shows the challenge in identifying such high-variant keyphrases. Despite the challenge, \text{\joint{}} and its informative ranking version (Rank) significantly outperform other methods in predicting non-entity keyphrases. The results reveal that capturing the global informativeness of keyphrase candidates using learning-to-rank helps overcome some difficulty of non-entity keyphrase extraction.

\subsection{Case Studies}

Table~\ref{tab:cases} exhibits some extraction cases of long and non-entity keyphrases from \text{\joint{}} and its two ablated versions for the OpenKP dataset. 

The first two cases show the long-keyphrase extraction scenarios. In the first case, the bi-gram keyphrase ``walter lagraves'' has been successfully predicted by all three methods, but the longer keyphrase ``mosquito control board race'' has only been extracted by \text{\joint{}}. Besides, \text{\joint{}} successfully predicted the longer keyphrase ``surplus lines export eligibility'' in the second case, while its two ablated versions only extracted the short one ``compliance corner''. 

The remaining cases are non-entity examples. In contrast to entity keyphrases, non-entity keyphrases have more variable forms and may appear less frequently in contexts of the given documents, e.g., entity ``fire port'' and non-entity ``dv to laptop'' in the third case, and entity ``north country realty'' and non-entity ``the beauty of the north'' in the fourth case. All three methods have successfully predicted the entity keyphrases, but the non-entity ones were only extracted by the full version of \joint{}. 
\section{Conclusion}
This paper presents \text{JointKPE}, a multi-task architecture built upon pre-trained language models for open-domain KPE, which can capture both local phraseness and global informativeness when extracting keyphrases. Our experiments demonstrate \text{JointKPE} 's effectiveness on both open-domain and scientific scenarios and across different pre-trained models. Comprehensive empirical studies further indicate that \text{JointKPE} can alleviate the problem of preferring shorter and entity-style keyphrases in previous neural KPE methods and exhibits more balanced performance on keyphrases of different lengths and various types.

%

\section*{Acknowledgments}
This work is partly supported by National Natural Science Foundation of China (NSFC) under grant No. 61872074 and 61772122; Beijing National Research Center for Information Science and Technology (BNR2019ZS01005).

\bibliographystyle{splncs04}
\bibliography{bibliography}

\end{document}